%% file: elope_arxiv.tex
\documentclass[10pt,twocolumn,letterpaper]{article}

\usepackage{wacv}
\usepackage{times}
\usepackage{epsfig}
\usepackage{graphicx}
\usepackage{amsmath}
\usepackage{amssymb}

\wacvfinalcopy 

\ifwacvfinal\fi
\begin{document}

\title{ELoPE: Fine-Grained Visual Classification with Efficient Localization, Pooling and Embedding}

\author{Harald Hanselmann and Hermann Ney \\
Human Language Technology and Pattern Recognition Group\\
RWTH Aachen University\\
Aachen, Germany\\
{\tt\small <surname>@cs.rwth-aachen.de}
}

\maketitle
\ifwacvfinal\thispagestyle{empty}\fi

\begin{abstract}
The task of fine-grained visual classification (FGVC) deals with classification problems that display a small inter-class variance such as distinguishing between different bird species or car models. State-of-the-art approaches typically tackle this problem by integrating an elaborate attention mechanism or (part-) localization method into a standard convolutional neural network (CNN). Also in this work the aim is to enhance the performance of a backbone CNN such as ResNet by including three efficient and lightweight components specifically designed for FGVC. This is achieved by using global k-max pooling, a discriminative embedding layer trained by optimizing class means and an efficient bounding box estimator that only needs class labels for training. The resulting model achieves new best state-of-the-art recognition accuracies on the Stanford cars and FGVC-Aircraft datasets.
\end{abstract}

\section{Introduction}
Fine-grained visual classification (FGVC) refers to classification tasks where the differences between the different categories are very subtle. Examples of such tasks are the classification of bird species or differentiating between different car models. The general appearance of the categories is very similar (\eg all birds have two wings and a beak, cars typically have four wheels) and as result the inter-class variation is small. On the other hand, the intra-class variation can be quite high (\eg due to different poses). This makes FGVC a very challenging problem that receives a lot of attention in the research community. State-of-the-art approaches typically involve a backbone CNN such as ResNet \cite{he2016deep} or VGG \cite{simonyan2014very} that is extended by a method that localizes and attends to specific discriminative regions. These methods can become quite complex and sometimes require multiple passes through the backbone CNN.

In this work we aim to improve the performance of a given backbone CNN with little increase in complexity and requiring just a single pass through the backbone network. Specifically, we propose the following three steps:
\begin{itemize}
  \item Global k-max pooling: For FGVC-models, the final convolutional layer often still has a spatial resolution of $I \times J$ (e.g. for a ResNet-50 with $448 \times 448$ input the resolution is $14 \times 14$). A single feature vector describing the image can then be obtained by using global average or global max pooling. However, to approximate part-based recognition, we propose to use global k-max pooling, where the average over the $k$ maximal features is computed.
  \item Embedding layer: In a typical setup for face verification tasks, the test subjects (\ie classes) are not known during training, which means a standard softmax classifier can not be trained. CNNs are therefore often used to train a discriminative embedding space in which face images can be compared efficiently and accurately. The embeddings are learned using specifically designed loss functions such as center loss \cite{wen2016discriminative}, triplet loss \cite{schroff2015facenet} or DFF \cite{hanselmann2017deep}. We insert such an embedding layer trained with a loss function based similar to \cite{hanselmann2017deep} into the backbone CNN as penultimate layer. We show that this greatly improves the performance of the softmax classifier.
  \item Localization module: Using bounding boxes to crop the input images typically improves the performance of the classification model. In order to avoid having to rely on human bounding box annotations we train an efficient bounding box detector that can be applied before the image is processed by the backbone CNN. This localization module is lightweight and trained using only the class labels. Bounding box annotations are not needed.
\end{itemize}

We evaluate our model on three popular FGVC datasets from different domains. The first dataset is CUB200-2011 \cite{birds} where the task is the classification of bird species. The second dataset is Stanford cars \cite{cars} where different car models are classified and the third is FGVC-Aircraft \cite{aircraft} for the classification of different aircraft models. We obtain very competitive results on all three datasets and to the best of our knowledge new best state-of-the-art results for the latter two.

\subsection{Related work}
As mentioned in the introduction FGVC has received a lot of attention in the research community. As a result, many different approaches have been proposed. Especially using some form of visual attention has been very popular lately \cite{zheng2017learning,zhao2017diversified,yang2018learning,fu2017look,wu2018deep,li2017dynamic,sun2018multi}. The work presented in \cite{sun2018multi} is of particular relevance since here also an embedding loss is used. However this loss is not used to train an independent embedding layer but to regulate the defined attention mechanism.

Spatial transformations that extract the discriminative parts of the input can also be seen as a form of attention. For example, the well known spatial transformer introduced in \cite{jaderberg2015spatial} is capable of learning global transformations (\eg affine transformations), but is known to be difficult to train and usually needs a second large network to estimate the transformation parameters. In \cite{xu2018attend} a module to learn pixel-wise translations is proposed. However, this module is applied very late in the network, possibly due to being reliant on high-level features. As a result only the last layers can profit from the localized input. In \cite{simonelli2018increasingly} an ensemble of networks is learned sequentially, where each network is trained based on a spatial transformation derived from the previous network. This means that each input image needs to be passed through multiple networks. Our localization module fits into the category of spatial transformations (limited to scale and translation), but it is very lightweight and easy to train while still being able to significantly boost the recognition performance.

The work presented in \cite{liang2018gmm} proposes to train a gaussian mixture model based on part proposals provided by selective search. However, this requires a looped training procedure with the EM-algorithm.

Another popular approach is based on bilinear models \cite{lin2015bilinear} which can lead to issues with efficiency due to very high dimensional features and multi-stream architectures. Also other second-order pooling methods such as the work in \cite{li2018towards} often results in very high dimensional features.

Other approaches include learning global and patch features in an asymmetric multi-stream architecture \cite{wang2018learning}, learning a complex sequence of data augmentation steps from the data \cite{cubuk2018autoaugment}, deep layer aggregation \cite{yu2018deep} or the training of very large networks (\eg 557 million parameters) \cite{huang2018gpipe}. It is also possible to boost the performance by obtaining more training data \cite{cui2018large,krause2016unreasonable}.

\section{Overview}
\begin{figure*}[t]
\begin{center}
   \includegraphics[width=0.65\linewidth]{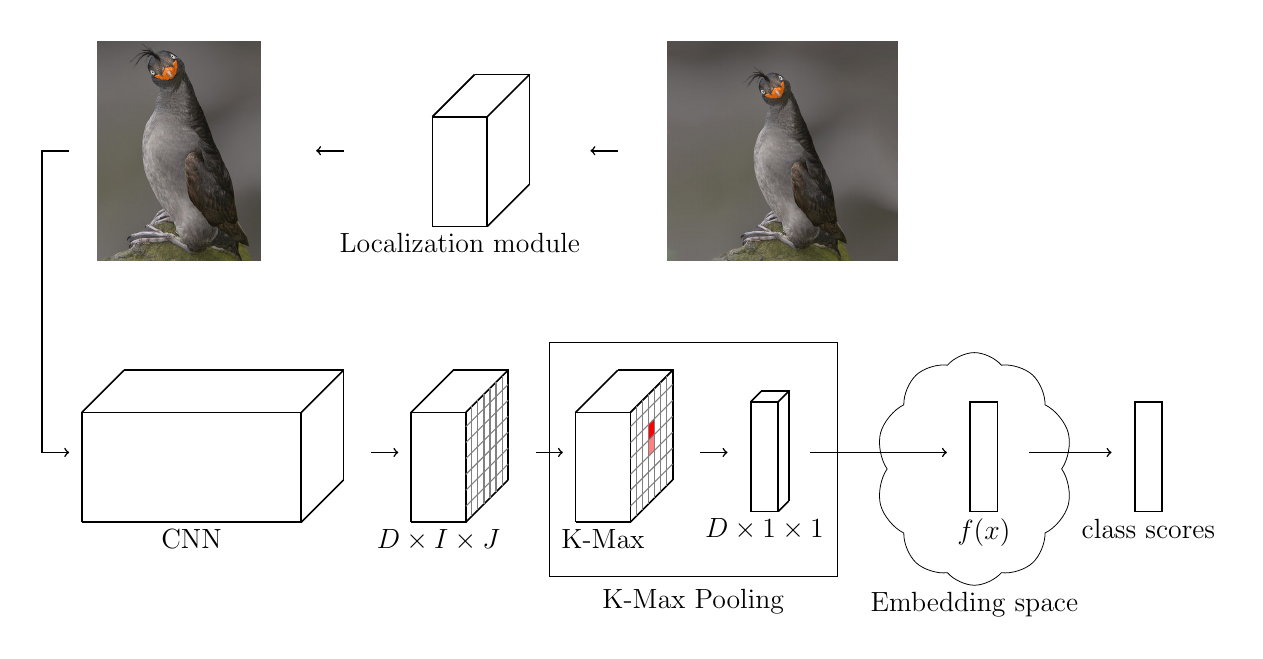}
\end{center}
   \caption{Overview of the proposed model including a lightweight localization module, global k-max pooling and an embedding layer.}
\label{fig:overview}
\end{figure*}

The goal of this work is to improve the performance of a given backbone CNN (\eg a ResNet \cite{he2016deep}) with lightweight components that do not require multiple passes through the backbone CNN or significantly higher runtime or memory usage during testing. We achieve this goal by adding three components, a localization module, global k-max pooling and an embedding layer. An overview of the resulting model in the testing stage is given in Figure \ref{fig:overview}. The input image that needs to be classified is forwarded through the localization module which estimates the bounding box of the object in the image and returns a cropped image. This cropped image is then forwarded thought the backbone CNN that contains global k-max pooling and the embedding layer at the later stages. The classification result is then given by a softmax classification layer.

In the training stage the model is trained jointly with a standard cross-entropy loss $L_{CE}$ applied at the classification layer and a specific loss function $L_e$ that is applied at the embeddings layer:

\begin{align}
  \label{eq:loss}
  L &= L_{CE} + \lambda L_{e}
\end{align}

The localization module is trained separate training step (c.f. Section \ref{section:loc}).

\section{Global K-Max Pooling}
\label{section:gkmp}
Often global max pooling (GMP) or global average pooling (GAP) is used between the last convolutional layer and the classification layer of a CNN. These pooling operations allow to break down the spatial dimension of the final convolutional layer and obtain a single vector describing the image. For FGVC it has been shown that part-based approaches (\eg \cite{zheng2017learning}) can boost the classification performance. For this reason we propose to use global k-max pooling (GKMP). This two-step pooling procedure first applies k-max pooling \cite{koniusz2013comparison} at the last convolutional layer which is followed by an averaging operation over the $K$ selected maximal values in each feature map. This way the network can learn features that activate at the $K$ most important parts of the image (during back-propagation the error gets propagated through the $K$ most important parts instead of just one as with GMP or all of them as with GAP).
This could be seen as a very simple form of attention.

The global k-max pooling layer can be defined as follows. Given an input image $x$, let $y \in \mathbb{R}^{D \times I \times J}$ be the output of the last convolutional layer of a CNN, where $y$ has a spatial resolution of $I \times J$ (in this work typically $14 \times 14$) and contains $D$ feature maps. Further, given a specific $d \in \{1,...,D\}$ the sorted vector $S_d$ contains the values at $y_d$ sorted in descending order:
\begin{align}
  \label{eq:sortedmax}
  S_d = sorted\_desc( \{ y_{d,i,j} | &i \in \{1,...,I\}, \nonumber \\
        &j \in \{1,...,J\} \})
\end{align}
The global k-max pooling operation for a specified $K$ is then is then defined as:
\begin{align}
  \label{eq:kmaxpooling}
  GKMP(y)_d &= \frac{1}{K} \sum_{k=1}^K S_{d,k}
\end{align}
This definition corresponds to the pooling operation used in \cite{durand2016weldon}, except in \cite{durand2016weldon} also the $K$ minimal activations are included. However, preliminary experiments suggested that including the minimal activations does not help the recognition performance. Therefore we use global k-max pooling as defined in Equation (\ref{eq:kmaxpooling}).

Note that if we select $K=1$ then Equation (\ref{eq:kmaxpooling}) results in the standard GMP, while a choice of $K=I\cdot J$ leads to GAP. In this work we chose $K=4$ in all experiments.

\subsection{Global k-max pooling with weighted averaging}
\label{section:gkmp_wavg}
GKMP can be extended by including weights in the averaging operation in Equation (\ref{eq:kmaxpooling}) similar to \cite{weng2013learning}:
\begin{align}
  \label{eq:kmaxpooling_wavg}
  GKMP(y)_d &= \frac{1}{K} \sum_{k=1}^K w_k \cdot S_{d,k}
\end{align}
This formulation extends the network with only $K$ parameters that regulate the contribution of the $K$ maximal activations in each feature map.

\section{Embedding layer}
\label{section:emb}

The embedding layer is inserted between GKMP and the classification layer. The idea is to map the images into an discriminative embedding space, where the distances between images of the same class are small, while the distances between images of different classes are large. This concept is known from face verification \cite{schroff2015facenet,wen2016discriminative,hanselmann2017deep} and metric learning \cite{rippel2015metric}. Different from those two tasks, we do not compare images directly in the embedding space, but use it as an intermediate layer (more specifically as penultimate layer). The classification is done by a standard classification layer trained on the embedding space with cross-entropy. We argue that one of the advantages of this approach is that it can help mitigate the issue of limited training data that we often see in FGVC tasks (see Table \ref{table:data}).

The embedding space is trained using a specific loss function $L_e$ applied directly to output of this intermediate layer. In this work we use a formulation based on optimizing class means \cite{wen2016discriminative,hanselmann2017deep} since this is easy to integrate into the training and does not require any specific batch construction schemes (unlike tuplet-based losses such as the triplet loss \cite{schroff2015facenet} or the NPair loss \cite{sohn2016improved}). For each class $c$ a feature vector is computed (and updated online during training) that describes the class mean $\mu_c$ within the embedding space. The goal of the loss function $L_e$ is to minimize the distance of each image within a batch to its respective class mean, while maximizing the distance between means of different classes. 

The loss function is composed of two parts:
\begin{align}
  \label{eq:centerLossCE}
  L_e &= L_{w} + L_{b}
\end{align}
The first part $L_{w}$ is the within-class (intra-class) loss that minimizes the distances of the images to their class means, while the second part $L_{b}$ is the between-class (inter-class) loss that maximizes the distances between class means.

\subsection{Within-class loss}
\label{section:intraLoss}
We use the same formulation for the within-class loss as in \cite{hanselmann2017deep} (which is derived from the center-loss proposed in \cite{wen2016discriminative}) including an additional $l2$-normalization. Given classes $c \in \{1,...,C\}$ and a batch of images $x_n$ with $n \in \{1,...,N\}$ let $f(x_n)$ be the normalized output of the embedding layer. Assuming we are currently in training iteration $t$ the first step is to update the class means using the data points in the current batch and the class means from the previous iteration $t-1$
\begin{align}
  \label{eq:centerUpdate}
  \mu_c^t = \mu_c^{t-1} - \alpha \Delta \mu_c^{t-1}
\end{align}
where the hyper-parameter $\alpha$ can be considered the learning rate of the class means. The term $\Delta \mu_c^{t-1}$ is defined as
\begin{align}
  \Delta \mu_c^{t-1} &= \frac{ \sum\limits_{n=1}^N \delta(c_n,c) (\mu_{c_n}^{t-1} - f(x_n)) } { 1 + \sum\limits_{n=1}^N \delta(c_n,c) }
\end{align}
and $\delta(c_n,c)$ is the Kronecker-function:
\begin{align}
  \delta(c_n,c) &=  \begin{cases} 1  & c = c_n \\ 0 & c \neq c_n \end{cases}
\end{align}
Note that this update creates a functional dependence between the class means and their corresponding images within the batch which has to be considered during back-propagation \cite{hanselmann2017deep}.

The within-class loss function is then defined as
\begin{align}
  \label{eq:intraLoss}
  L_{w} &= \frac{1}{2N} \sum_{n=1}^N  \lVert f(x_n) - \mu_{c_n}^t \rVert_2^2
\end{align}

\subsection{Between-class loss}
\label{section:interLoss}
The between-class loss maximizes the distances between the updated class means:
\begin{align}
  \label{eq:interLoss}
  L_{b} =& \frac{\gamma}{4 \vert P \vert} \sum_{(k,c) \in P}  \text{max} \left ( m - \lVert \mu_k^t - \mu_c^t \rVert_2^2, 0 \right )^2
\end{align}
The terms $\mu_k^t$ and $\mu_c^t$ are the new class means updated with the features $f(x_n)$ from the current batch as computed in Equation (\ref{eq:centerUpdate}). The margin $m$ defines a threshold for the distances to be penalized and $\gamma$ controls the contribution of the between-class part to the embedding loss $L_e$. The set $P$ with cardinality $\vert P \vert$ contains all class-pairs in the current batch. 

To see why squaring the maximum in Equation (\ref{eq:interLoss}) is important we have a look at the gradient with respect to $f(x_n)$:

\begin{align}
  \label{eq:gradienticdbase}
  &\frac{\partial L_{b}}{\partial f(x_n)} = \nonumber\\
  &\sum \limits_{  (k,c) \in P } \begin{cases} 0  & m \le \lVert \mu_k^t - \mu_c^t \rVert_2^2 \\ grad(m,\mu_k^t,\mu_c^t)  & \text{otherwise} \end{cases}
\end{align}
with
\begin{align}
  \label{eq:gr}
  &grad(m,\mu_k^t,\mu_c^t) = \nonumber\\
  &\underbrace{(m - \lVert \mu_k^t - \mu_c^t \rVert_2^2)}_{d(m,\mu_k^t,\mu_c^t)} \frac{\partial}{\partial f(x_n)} \left ( \frac{\gamma}{2 \vert P \vert} \lVert \mu_k^t - \mu_c^t \rVert_2^2 \right )
\end{align}
and

The squared maximization leads to the appearance of the distance $d(m,\mu_k^t,\mu_c^t)$ in the gradient. As a result the gradient gets larger if the distance between two class means gets smaller which encourages the model to focus more on improving the distance of class-pairs with very close means. This should help to reduce classification errors due to the confusion of images from class-pairs with close class means.

While the formulation of the between-class loss defined above is close to \cite{hanselmann2017deep}, the appearance of $d(m,\mu_k^t,\mu_c^t)$ in the gradient is a key difference, since in \cite{hanselmann2017deep} the maximization in the between-class loss is not squared. In addition there are two more differences. First the centers are based on $l2$-normalized features, which makes the choice for the margin $m$ easier, since distances between $l2$-normalized vectors are restricted by a certain range. 
The second difference is that in \cite{hanselmann2017deep} the set $P$ contains a sampled subset of all class-pairs in the current batch. Since we work with much smaller batch sizes (see Section \ref{section:eval}) compared to \cite{hanselmann2017deep}, we use all pairs within a batch.

The embedding layer as defined above is also closely related to the attention regulation proposed in OSME-MAMC \cite{sun2018multi}. The latter uses a metric learning loss to learn correlations between the output of different attention branches. If the multiple attention branches were replaced by a single fully connected layer then this method would be equivalent to training an embedding layer with the NPair loss \cite{sohn2016improved}.

\section{Localization module}
\label{section:loc}
It can be observed that cropping the images based on bounding boxes of the objects that need to be recognized can improve the classification performance. Since we do not want to use any annotations apart from the class labels in training (and of course no annotations at all during testing) we need a way to obtain these localizations without any additional annotations if we want to profit from the increased performance the localizations can provide. 
We design such a localization module based on the following observations.
\begin{figure}[t]
\begin{center}
  \begin{tabular} {c c c c c}
    \includegraphics[width=0.13\linewidth]{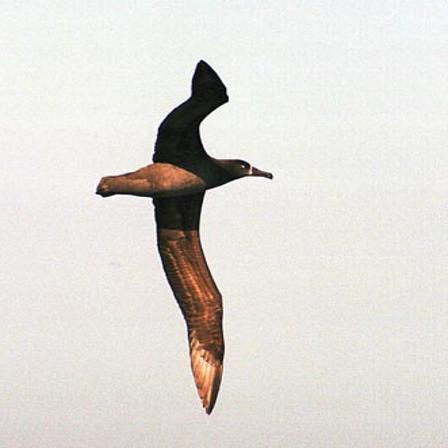} &
    \hspace{2mm}
    \includegraphics[width=0.13\linewidth]{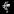} &
    \includegraphics[width=0.13\linewidth]{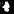} &
    \includegraphics[width=0.13\linewidth]{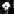} \hspace {2mm} &
    \includegraphics[width=0.13\linewidth]{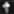} \\
    \includegraphics[width=0.13\linewidth]{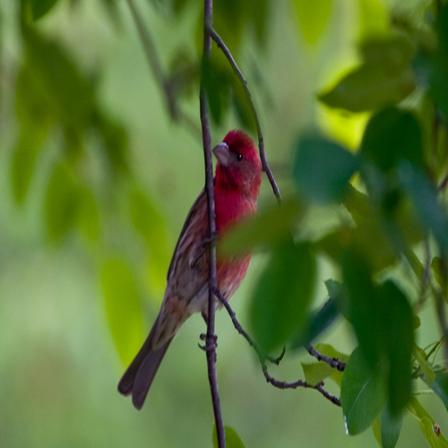} &
    \hspace{2mm}
    \includegraphics[width=0.13\linewidth]{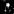} &
    \includegraphics[width=0.13\linewidth]{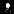} &
    \includegraphics[width=0.13\linewidth]{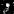} \hspace {2mm} &
    \includegraphics[width=0.13\linewidth]{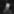}
  \end{tabular}
\end{center}
\caption{Min-max normalized activations of the last convolutional layer of a trained model. The three images in the middle are the activations in a specific feature map, while the last image shows the mean over all feature maps.}
\label{fig:activations}
\end{figure}

The final convolutional layer of a trained classification model typically contains higher activations in positions corresponding the discriminative areas of the image (see Figure \ref{fig:activations}). By computing the mean over all feature maps a quite accurate heat map can be computed for the object in the image. This heat map can the be used to find the boundaries of the object within the final layer. A similar observation has been made in \cite{zhou2016learning}, but in contrast to \cite{zhou2016learning} we do not consider class-specific heat maps due to the small inter-class variations in FGVC. 

The downside with this approach is that we obtain these bounding boxes only at the final layer while the full image needs to be forwarded through all previous layers. One option would be to apply ROI-Pooling after the final layer based on the estimated bounding box. However, this would mean that only the fully connected layers following the last convolutional layer would profit from focusing on the discriminative area of the image. All other layers still have to operate on the full image and we can not fully exploit the potential from using the bounding boxes (see Figure \ref{plot:bb}). An alternative would be to pass the image through the full backbone network, extract the bounding boxes and then train a second, more precise model based on the bounding boxes (similar to \cite{simonelli2018increasingly}). However, with this approach the image would have to be passed through a full network twice, one pass to obtain the bounding box and a second pass for classification. This is not very efficient. To avoid such a multi-pass procedure we propose to train a very lightweight localization module that predicts the bounding boxes and is integrated into the backbone network such that an image can be processed in one pass.

\begin{figure}[t]
\begin{center}
   \includegraphics[width=0.97\linewidth]{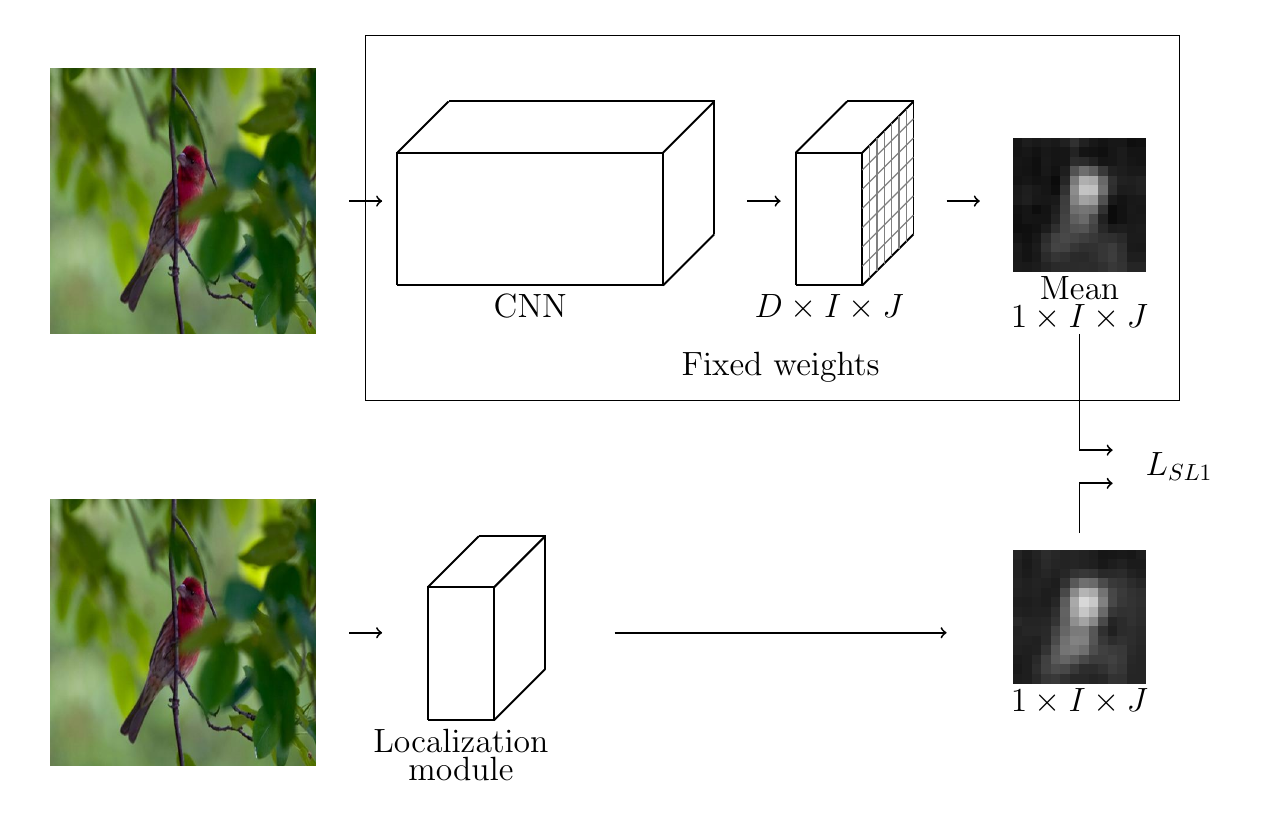}
\end{center}
   \caption{Training procedure for the localization module: The localization module learns to directly predict the heat maps by minimizing the mean squared error with the heat maps generated by the backbone CNN.}
\label{fig:loc_training}
\end{figure}

The architecture of the localization module is equivalent to the first few layers of a ResNet-50 (initialized from the trained classification model) until the end of the first residual block including an initial down-sampling layer that resizes the input to a spatial resolution of $64 \times 64$. The module has only $220000$ parameters and can be added to the classification model without taking up much runtime or memory. The output is of the size $1 \times I \times J$, the same as the mean of the last convolutional layer of the trained classification model. The localization module is trained by feeding an input image through the trained classification model and the localization module. The outputs are compared using the smooth L1 loss \cite{girshick2015fast} (see Figure \ref{fig:loc_training}). During back-propagation the weights of the trained classification model are fixed and only the localization module is trained. This way the localization module learns to directly predict the heat maps.

\begin{figure}[t]
\begin{center}
  \begin{tabular} {c c c}
    \includegraphics[width=0.15\linewidth]{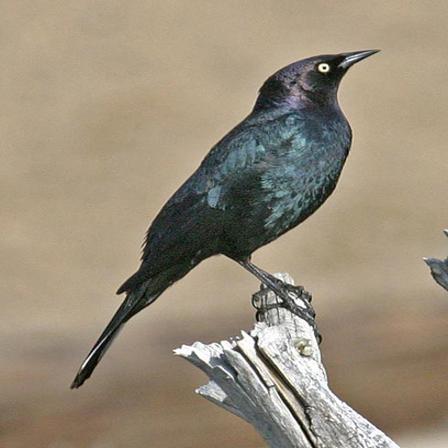} &
    \includegraphics[width=0.15\linewidth]{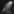} &
    \includegraphics[width=0.15\linewidth]{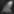} \\
    \includegraphics[width=0.15\linewidth]{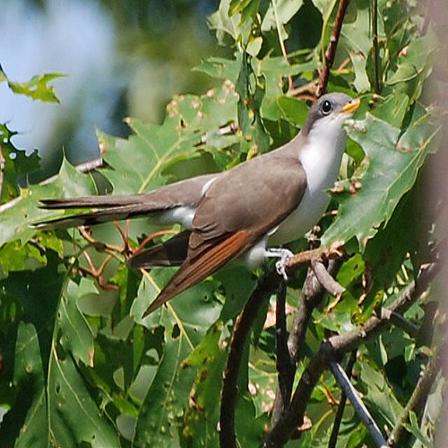} &
    \includegraphics[width=0.15\linewidth]{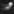} &
    \includegraphics[width=0.15\linewidth]{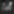} \\
  \end{tabular}
\end{center}
\caption{True mean (middle) of the trained classification model compared to estimated mean (right).}
\label{fig:loc_examples}
\end{figure}

Examples of the estimated heat maps are given in Figure \ref{fig:loc_examples}. The localization module is able to predict the heat maps very accurately, even though less focused on a specific part of the bird (especially in the second and third row).

To obtain the final bounding box we we process the heat map using min-max normalization and binarization based on a given threshold $\tau$. The bounding box is then the smallest rectangle containing all pixels where the heat map has a value that is greater than $\tau$.

\section{Experimental evaluation}
\label{section:eval}
\begin{table}
  \begin{center}
  \begin{tabular} {|l|c|c|c|}
    \hline
    Name & \#Train & \#Test & \#Classes\\
         & images  & images & \\
    \hline
    \hline
    CUB200-2011 \cite{birds} & 5994 & 5794 & 200 \\
    \hline
    Stanford cars \cite{cars} & 8144 & 8041 & 196\\
    \hline
    FGVC Aircraft \cite{aircraft} & 6667 & 3333 & 100\\
    \hline
  \end{tabular}
  \end{center}
  \caption{Datasets used for the evaluation.}
  \label{table:data}
\end{table}
We evaluate our proposed approach on three popular datasets for fine-grained classification, CUB200-2011 \cite{birds}  for bird-species classification, Stanford cars \cite{cars} for the classification of car models and FGVC-Aircraft \cite{aircraft} for the classification of airplane models. The statistics of the three datasets are given in Table \ref{table:data}. On all datasets we use only the class labels annotations. We do not use any additional annotations such as bounding boxes or part annotations.

As backbone CNN we select ResNet \cite{he2016deep} where we try the two variants ResNet-50 and ResNet-101 (the analysis in Section \ref{section:loc_exp} and \ref{section:ablation} is done using ResNet-50 as backbone CNN.). The models are pretrained on the ImageNet \cite{deng2009imagenet} from which we remove all images that overlap with the test sets of the three FGVC tasks used for this evaluation.

Since our added components are all lightweight and do not occupy much memory we are able to train both ResNet variants on a single NVIDIA GeForce$^{\textregistered}$ GTX 1080 Ti GPU with 11GB memory with a batch-size of $14$ and input images with the spatial resolution $448 \times 448$. We use this batch-size and input resolution for all experiments. The models are trained with standard back-propagation for $90$ epochs with momentum of $0.9$ and weight decay of $0.001$. The starting learning rate is $0.003$ which is reduced by a factor of $10$ after $30$ epochs. The weighted average pooling (see Section \ref{section:gkmp_wavg}) is applied as finetuning step for $30$ more epochs. The approach is implemented using the torch7 framework \cite{collobert2011torch7}.

There is a number of hyper-parameters to set for our approach. We perform a very limited search for the hyper-parameters on CUB200-2011 using a validation set separated from the training images. The parameters are quite robust and we can use the same set of hyper-parameters in all other experiments. The threshold $\tau$ for the localization module is the only exception, where we use a slightly smaller value on the Stanford cars and the FGVC-Aircraft dataset. The exact values for the hyper-parameters are given in Table \ref{table:params}.

In Section \ref{section:birds}, \ref{section:cars} and \ref{section:aircraft} we compare to state-of-the-art approaches using a similar experimental setting, specifically with the same training data (ImageNet and training set of the FGVC task at hand) and no bounding box or part annotations. Note that for some datasets better results can be achieved by acquiring large amounts of additional training data \cite{cui2018large,krause2016unreasonable}. 

\begin{table}
  \begin{center}
  \begin{tabular} {|c|c|c|}
    \hline
    Hyper-    & Reference & Value \\
    parameter &           & \\
    \hline
    \hline
    $\alpha$    & Equation (\ref{eq:centerUpdate}) & 0.5 \\
    $\lambda$   & Equation (\ref{eq:loss}) & 2.0\\
    $\gamma$   & Equation (\ref{eq:interLoss}) & 16.0 \\
    $m$        & Equation (\ref{eq:interLoss}) & 0.75 \\
    $K$        & Equation (\ref{eq:kmaxpooling}) &  4\\
    $\tau$     & Section (\ref{section:loc})& 0.3/0.2\\
    \hline
  \end{tabular}
  \end{center}
  \caption{Hyper-parameters used in the experiments. The threshold $\tau$ is 0.3 on CUB200-2011 and 0.2 on the other two datasets.}
  \label{table:params}
\end{table}

\subsection{Computational efficiency}
\begin{table}
  \begin{center}
  \begin{tabular} {|c|c|c|}
    \hline
    Method & FLOPs & Parameters  \\
    \hline
    \hline
    Baseline  & $1.648 \times 10^{10}$ & 24M \\
    Ours      &  $1.654 \times 10^{10}$ & 25M \\
    \hline
  \end{tabular}
  \end{center}
  \caption{Computational efficiency comparison between baseline ResNet-50 (input 1x3x448x448) and our approach.}
  \label{table:complex}
\end{table}
A comparison of the computational efficiency between a baseline ResNet-50 and our approach with ResNet-50 as backbone is given in Table \ref{table:complex}. With the typical input resolution of $448 \times 448$ the additional complexity in terms of FLOPs (multiply-adds) caused by our approach is very small (one reason is that the localization module operates on low resolution input ($64 \times 64$)). There is also no large increase in parameters highlighting the efficiency of the proposed approach.

\subsection{Localization module}
\label{section:loc_exp}
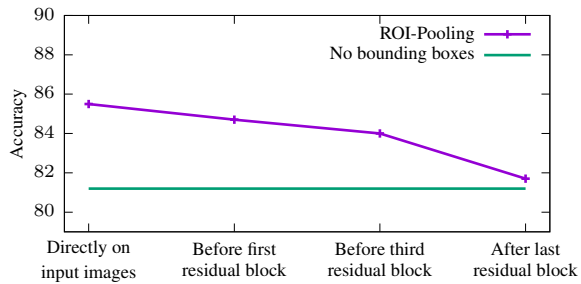
\begin{figure}[t]
  \scriptsize
  \begin{center}
    \input{bb.tex}
  \end{center}
  \caption{Accuracies for ROI-Pooling with ground-truth bounding boxes applied at different depths of a CNN on CUB200-2011 (ResNet-50 with GAP and no embedding learning).}
  \label{plot:bb}
\end{figure}

In Figure \ref{plot:bb} we analyze the effect on the recognition performance if the image or feature maps are cropped based on ground-truth bounding boxes (ROI-Pooling). We can observe that the earlier the ROI-Pooling is applied the better is the recognition accuracy, confirming our expectation in Section \ref{section:loc}.

\begin{table}
  \begin{center}
  \begin{tabular} {|c|c|}
    \hline
    Method    & Accuracy[\%]\\
    \hline
    \hline
    GoogleNet-GAP \cite{zhou2016learning} & 41.0 \\
    ResNet-50 mean feature map & 58.6 \\  
    Localization module & 68.9 \\ 
    \hline
  \end{tabular}
  \end{center}
  \caption{Accuracy of localization if intersection over union (IoU) is at least 0.5.}
  \label{table:iou}
\end{table}

The accuracy of the bounding boxes can be calculated by counting a bounding box as correct if the intersection over union (IoU) with the ground truth bounding box is at least 0.5. We can observe in Table \ref{table:iou} that using the mean over baseline ResNet-50 feature maps of the last convolutional layer already achieves a better accuracy then what is reported in \cite{zhou2016learning}. However, on top of being much more efficient the localization module is also even more accurate. We argue that this is due to the slightly more general heatmaps generated by the localization module as a result of the approximation (see Figure \ref{fig:loc_examples}).

\begin{figure}[t]
\begin{center}
  \begin{tabular} {c c c}
    \includegraphics[width=0.27\linewidth]{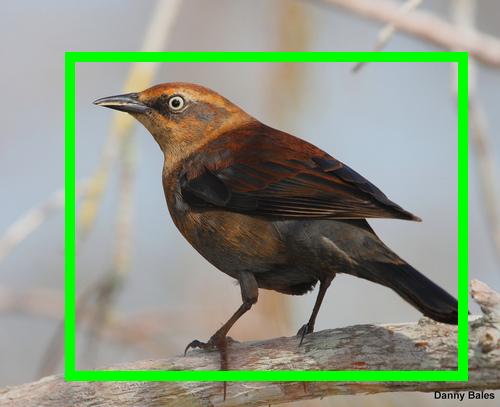} &
    \includegraphics[width=0.27\linewidth]{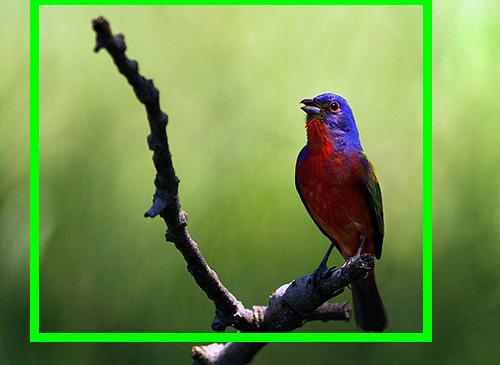} &
    \includegraphics[width=0.27\linewidth]{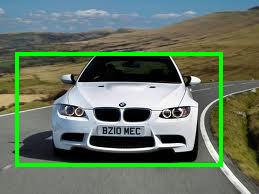} \\
    \includegraphics[width=0.27\linewidth]{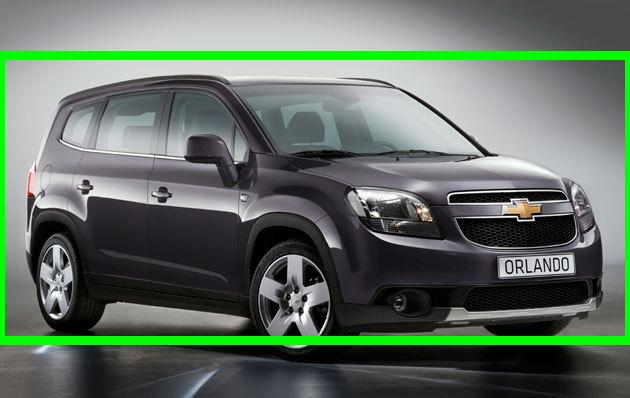} &
    \includegraphics[width=0.27\linewidth]{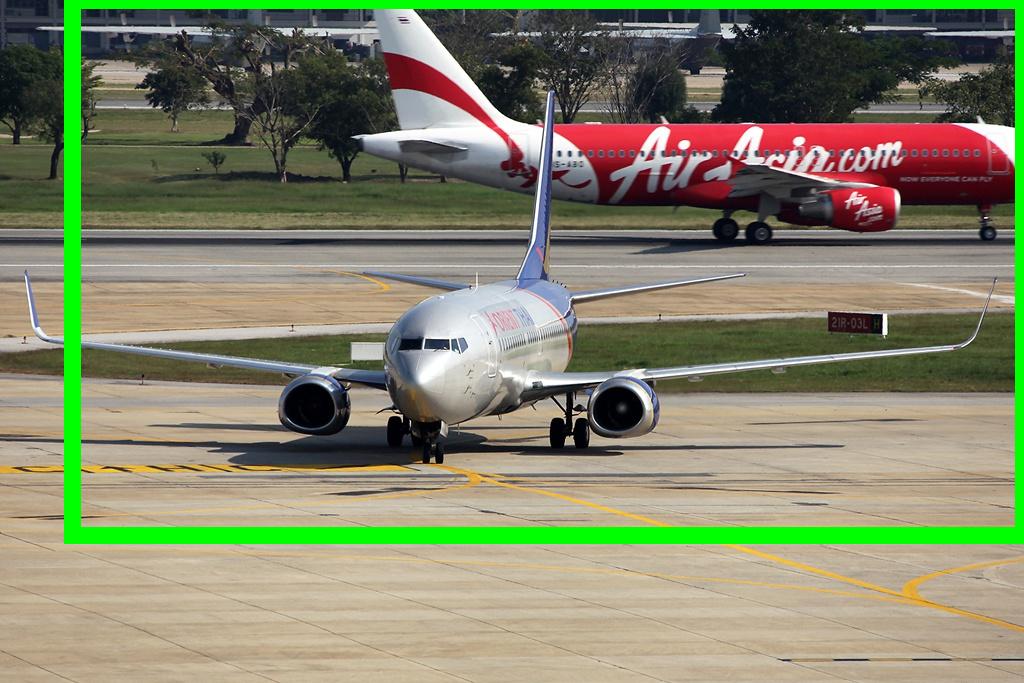} &
    \includegraphics[width=0.27\linewidth]{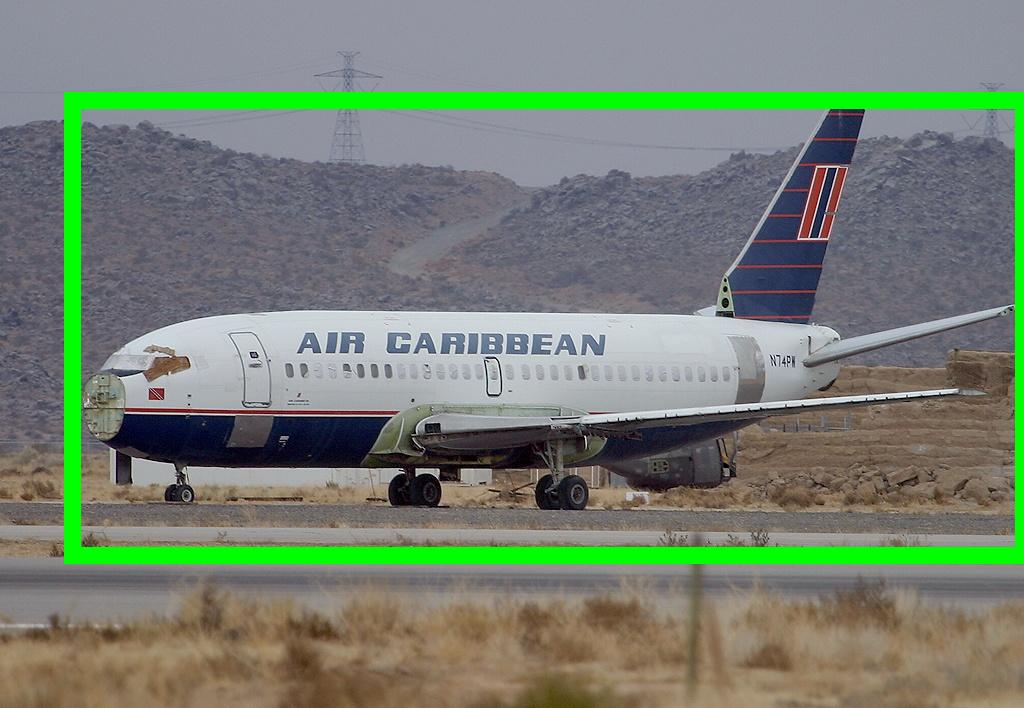} 
  \end{tabular}
\end{center}
\caption{Examples of bounding boxes detected by the localization module.}
\label{fig:bb_examples}
\end{figure}

In Figure \ref{fig:bb_examples} we show qualitative results of the bounding boxes estimated for validation images by the localization module. We can observe that the localization module is able to find quite accurate bounding boxes. In some cases (right column of Figure \ref{fig:bb_examples}) the localization module over-estimates the bounding box due to other objects in the image such as a branch or a distracting background. Another interesting reason for over-estimated bounding boxes can be multiple instances of the given FGVC domain such as two different airplane models in the same image. However, in each of these cases the classifier still has a good chance of finding the correct class, since the objects are still present in the cropped image.

\subsection{Ablation study}
\label{section:ablation}
Table \ref{table:analysis} shows how the components contribute to the recognition performance. The two baseline results with GAP and GMP include a fully connected layer with dimension $512$ before the classification layer. We can observe in Table \ref{table:analysis} that adding an embedding loss $L_e$ and adding the localization module leads to the largest boost in performance at about $2\%$ absolute ($86.9\%$, using ground-truth bounding boxes yields $87.5\%$). The addition of global k-max pooling, weighted average finetuning and the full embedding loss compared to only the within-class part lead to an improvement of about $0.5\%$ each.

\begin{table*}[h]
  \begin{center}
  \begin{tabular} {|c|c|c|c|c|}
    \hline
    Pooling & $L_e$     & Localization & Weighted & Accuracy[\%] \\
          &             & module       & Average  & \\
    \hline
    \hline
    GAP   &             &              & & 81.2 \\
    GMP   &             &              & & 82.2 \\
    GKMP  &             &              & & 82.7\\
    GKMP  &             & \checkmark   & & 84.2\\
    GKMP  & Center-loss \cite{wen2016discriminative} & & & 84.1\\
    GKMP  & DFF \cite{hanselmann2017deep}  & & & 84.6\\
    GKMP  & $L_w$       &              & & 84.4\\
    GKMP  & $L_w + L_b$ &              & & 84.9\\
    GKMP  & $L_w + L_b$ & \checkmark   & & 86.9\\
    GKMP  & $L_w + L_b$ & \checkmark   & \checkmark & 87.4\\
    \hline
  \end{tabular}
  \end{center}
  \caption{Results on CUB200-2011 with a ResNet-50. The notation $L_e = L_w$ means the embedding layer is trained only with the within-class part of the embedding loss, while $L_e = L_w + L_b$ means the embedding layer is trained with the full embedding loss. Weighted average refers to Section \ref{section:gkmp_wavg}. For comparison we include results obtained with \cite{wen2016discriminative} and \cite{hanselmann2017deep}.}
  \label{table:analysis}
\end{table*}

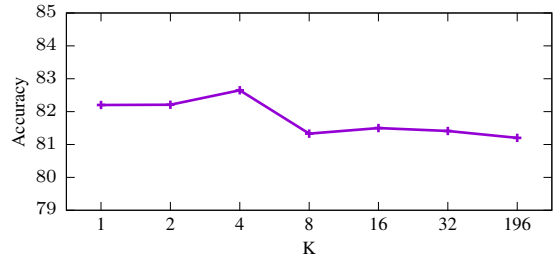
\begin{figure}[t]
  \scriptsize
  \begin{center}
    \input{kmax.tex}
  \end{center}
  \caption{Accuracies for different values of $K$ in $K$-max pooling on CUB200-2011.}
  \label{plot:kmax}
\end{figure}

The effect of the value selected for $K$ in GKMP is illustrated in Figure \ref{plot:kmax}. The special case $K=1$ is equivalent to GMP and $K=196$ is equivalent to GAP (the spatial dimension of the last convolutional layer is $14 \times 14$). We can observe that a value around $4$ seems to be optimal. The improvement compared to GMP is only $0.5\%$ absolute, but using a value larger than $1$ also enables us to apply the weighted averaging, which gives another performance boost (see Table \ref{table:analysis}).

\subsection{CUB200-2011}
\label{section:birds}
\begin{table}[h]
  \begin{center}
  \begin{tabular} {|l|c|c|}
    \hline
    Method & Backbone CNN & Accuracy[\%] \\
    \hline
    \hline
    STN \cite{jaderberg2015spatial} & BN-Inception & 84.1 \\
    MA-CNN \cite{zheng2017learning} & VGG-19 &  86.5 \\
    GMM \cite{liang2018gmm} & VGG-19 &  86.3 \\
    Spatial RNN \cite{wu2018deep} & M-Net/D-Net & 89.7\\
    Stacked LSTM \cite{Ge_2019_CVPR} & GoogleNet & \bf{90.4} \\
    DLA \cite{yu2018deep} & DLA-102 & 85.1 \\
    FAL \cite{xu2018attend} & ResNet-50 & 84.2 \\
    DT-RAM \cite{li2017dynamic} & ResNet-50 & 86.0 \\
    ISE \cite{simonelli2018increasingly} & ResNet-50 & 87.2 \\ 
    DFL-CNN \cite{wang2018learning} & ResNet-50 &  87.4 \\
    NTS-Net \cite{yang2018learning} & ResNet-50 &  87.5 \\
    DCL \cite{Chen_2019_CVPR} & ResNet-50 & 87.8 \\
    OSME-MAMC \cite{sun2018multi} & ResNet-101 & 86.5 \\
    iSQRT-COV \cite{li2018towards} & ResNet-101 & 88.7 \\
    \hline
    Ours   & ResNet-50  &  87.4 \\
    Ours   & ResNet-101 &  88.5 \\
    \hline
  \end{tabular}
  \end{center}
  \caption{Comparison of our approach with other state-of-the-art methods on CUB200-2011.}
  \label{table:birds}
\end{table}

In Table \ref{table:birds} we compare our approach to other state-of-the-art methods. In addition to the results reported in Table \ref{table:analysis} we also evaluate our method with ResNet-101 as backbone network. This includes all components (GKMP with weighted average pooling, embedding layer with full embedding loss, localization module). Our best result is with $88.5\%$ very competitive, even though a little less accurate then the best state-of-the-art result is reported in \cite{wu2018deep} and \cite{Ge_2019_CVPR}. However, these involve recurrent neural networks which can be computationally expensive. 

\subsection{Stanford cars}
\label{section:cars}
\begin{table}[h]
  \begin{center}
  \begin{tabular} {|l|c|c|}
    \hline
    Method & Backbone CNN &  Accuracy[\%] \\
    \hline
    \hline
    MA-CNN \cite{zheng2017learning} & VGG-19 &  92.8 \\
    GMM \cite{liang2018gmm} & VGG-19 &  93.5 \\
    DFL-CNN \cite{wang2018learning} & VGG-16  & 93.8 \\
    Spatial RNN \cite{wu2018deep} & M-Net/D-Net & 93.4\\
    DLA \cite{yu2018deep} & DLA-X-60-C  & 94.1 \\
    DT-RAM \cite{li2017dynamic} & ResNet-50 & 93.1 \\
    ISE \cite{simonelli2018increasingly} & ResNet-50 &  94.1 \\
    NTS-Net \cite{yang2018learning} & ResNet-50 & 93.9 \\
    DCL \cite{Chen_2019_CVPR} & ResNet-50 & 94.5 \\
    GPipe \cite{huang2018gpipe} & AmoebaNet-B &  94.8 \\
    AutoAugm \cite{cubuk2018autoaugment} & Inception-v4 &  94.8 \\
    OSME-MAMC \cite{sun2018multi} & ResNet-101 & 93.0 \\
    iSQRT-COV \cite{li2018towards} & ResNet-101 & 93.3 \\
    \hline
    Ours   & ResNet-50  & 94.5 \\
    Ours   & ResNet-101 &  \bf{95.0} \\
    \hline
  \end{tabular}
  \end{center}
  \caption{Comparison of our approach with other state-of-the-art methods on Stanford cars.}
  \label{table:cars}
\end{table}

The results on the Stanford cars dataset are given in Table \ref{table:cars}. Here we run the experiment with ResNet-50 and ResNet-101 with all components included. As mentioned earlier, we use the same hyper-parameters as for CUB200-2011 (apart from the threshold $\tau$). Again, our approach achieves a very competitive result. In fact, to the best of our knowledge, the accuracy of $95.0\%$ is the best result, even though by a small margin. 

\subsection{FGVC-Aircraft}
\label{section:aircraft}
\begin{table}[h]
  \begin{center}
  \begin{tabular} {|l|c|c|}
    \hline
    Method & Backbone CNN & Accuracy[\%] \\
    \hline
    \hline
    MA-CNN \cite{zheng2017learning} & VGG-19 &  89.9 \\
    GMM \cite{liang2018gmm} & VGG-19 &  90.5 \\
    DFL-CNN \cite{wang2018learning} & VGG-16 &  92.0 \\
    Spatial RNN \cite{wu2018deep} & M-Net/D-Net & 88.4\\
    DLA \cite{yu2018deep} & DLA-X-60 &  92.9 \\
    ISE \cite{simonelli2018increasingly} & ResNet-50 &  90.9 \\
    NTS-Net \cite{yang2018learning} & ResNet-50 & 91.4 \\
    DCL \cite{Chen_2019_CVPR} & ResNet-50 & 93.0 \\
    GPipe \cite{huang2018gpipe} & AmoebaNet-B & 92.9 \\
    AutoAugm \cite{cubuk2018autoaugment} & Inception-v4 & 92.7 \\
    iSQRT-COV \cite{li2018towards} & ResNet-101 & 91.4 \\
    \hline
    Ours   & ResNet-50  &  93.4 \\
    Ours   & ResNet-101 &  \bf{93.5} \\
    \hline
  \end{tabular}
  \end{center}
  \caption{Comparison of our approach with other state-of-the-art methods on FGVC-Aircraft.}
  \label{table:aircraft}
\end{table}

Similar to the Stanford cars dataset we use all components introduced in the previous sections and the same hyper-parameters as in Stanford cars. Also on FGVC-Aircraft (Table \ref{table:aircraft}) we can report a very competitive result which is again to the best of our knowledge the best state-of-the-art result.

\section{Conclusion}
In this work we presented three efficient methods to improve the classification performance of a backbone CNN for fine-grained visual classification. Specifically, we propose a lightweight localization module that relies only on class label annotations during training. We showed that even though the localization module can find reliable bounding boxes and significantly boost the recognition performance. Additionally we propose to use global k-max pooling to obtain a global vector describing the image. This approximates part-based modeling and can further be improved by learning weights to regulate the contribution of the maximal values in each feature map. Finally, we project the image descriptor into a discriminative embedding space from which the classification layer makes the classification. As an intermediate layer of the full classification network the embedding space is trained jointly with the full network and a specific loss function that optimizes class means. We evaluate our approach on three popular FGVC tasks and achieve competitive results on all three. In fact, on Stanford cars and FGVC-Aircraft we can report new best classification accuracies. 

\clearpage
{\small
\bibliographystyle{ieee}
\bibliography{elope_arxiv}
}

\end{document}

%% file: bb.tex
\begingroup
  \makeatletter
  \providecommand\color[2][]{%
    \GenericError{(gnuplot) \space\space\space\@spaces}{%
      Package color not loaded in conjunction with
      terminal option `colourtext'%
    }{See the gnuplot documentation for explanation.%
    }{Either use 'blacktext' in gnuplot or load the package
      color.sty in LaTeX.}%
    \renewcommand\color[2][]{}%
  }%
  \providecommand\includegraphics[2][]{%
    \GenericError{(gnuplot) \space\space\space\@spaces}{%
      Package graphicx or graphics not loaded%
    }{See the gnuplot documentation for explanation.%
    }{The gnuplot epslatex terminal needs graphicx.sty or graphics.sty.}%
    \renewcommand\includegraphics[2][]{}%
  }%
  \providecommand\rotatebox[2]{#2}%
  \@ifundefined{ifGPcolor}{%
    \newif\ifGPcolor
    \GPcolortrue
  }{}%
  \@ifundefined{ifGPblacktext}{%
    \newif\ifGPblacktext
    \GPblacktexttrue
  }{}%
  \let\gplgaddtomacro\g@addto@macro
  \gdef\gplbacktext{}%
  \gdef\gplfronttext{}%
  \makeatother
  \ifGPblacktext
    \def\colorrgb#1{}%
    \def\colorgray#1{}%
  \else
    \ifGPcolor
      \def\colorrgb#1{\color[rgb]{#1}}%
      \def\colorgray#1{\color[gray]{#1}}%
      \expandafter\def\csname LTw\endcsname{\color{white}}%
      \expandafter\def\csname LTb\endcsname{\color{black}}%
      \expandafter\def\csname LTa\endcsname{\color{black}}%
      \expandafter\def\csname LT0\endcsname{\color[rgb]{1,0,0}}%
      \expandafter\def\csname LT1\endcsname{\color[rgb]{0,1,0}}%
      \expandafter\def\csname LT2\endcsname{\color[rgb]{0,0,1}}%
      \expandafter\def\csname LT3\endcsname{\color[rgb]{1,0,1}}%
      \expandafter\def\csname LT4\endcsname{\color[rgb]{0,1,1}}%
      \expandafter\def\csname LT5\endcsname{\color[rgb]{1,1,0}}%
      \expandafter\def\csname LT6\endcsname{\color[rgb]{0,0,0}}%
      \expandafter\def\csname LT7\endcsname{\color[rgb]{1,0.3,0}}%
      \expandafter\def\csname LT8\endcsname{\color[rgb]{0.5,0.5,0.5}}%
    \else
      \def\colorrgb#1{\color{black}}%
      \def\colorgray#1{\color[gray]{#1}}%
      \expandafter\def\csname LTw\endcsname{\color{white}}%
      \expandafter\def\csname LTb\endcsname{\color{black}}%
      \expandafter\def\csname LTa\endcsname{\color{black}}%
      \expandafter\def\csname LT0\endcsname{\color{black}}%
      \expandafter\def\csname LT1\endcsname{\color{black}}%
      \expandafter\def\csname LT2\endcsname{\color{black}}%
      \expandafter\def\csname LT3\endcsname{\color{black}}%
      \expandafter\def\csname LT4\endcsname{\color{black}}%
      \expandafter\def\csname LT5\endcsname{\color{black}}%
      \expandafter\def\csname LT6\endcsname{\color{black}}%
      \expandafter\def\csname LT7\endcsname{\color{black}}%
      \expandafter\def\csname LT8\endcsname{\color{black}}%
    \fi
  \fi
    \setlength{\unitlength}{0.0500bp}%
    \ifx\gptboxheight\undefined%
      \newlength{\gptboxheight}%
      \newlength{\gptboxwidth}%
      \newsavebox{\gptboxtext}%
    \fi%
    \setlength{\fboxrule}{0.5pt}%
    \setlength{\fboxsep}{1pt}%
\begin{picture}(4320.00,2016.00)%
    \gplgaddtomacro\gplbacktext{%
      \csname LTb\endcsname%
      \put(372,388){\makebox(0,0)[r]{\strut{}$80$}}%
      \put(372,685){\makebox(0,0)[r]{\strut{}$82$}}%
      \put(372,982){\makebox(0,0)[r]{\strut{}$84$}}%
      \put(372,1279){\makebox(0,0)[r]{\strut{}$86$}}%
      \put(372,1575){\makebox(0,0)[r]{\strut{}$88$}}%
      \put(372,1872){\makebox(0,0)[r]{\strut{}$90$}}%
      \put(627,120){\makebox(0,0){\strut{}\shortstack{ \vspace{8pt} \\ Directly on \\ input images}}}%
      \put(1725,120){\makebox(0,0){\strut{}\shortstack{ \vspace{8pt} \\ Before first \\ residual block}}}%
      \put(2822,120){\makebox(0,0){\strut{}\shortstack{ \vspace{8pt} \\ Before third \\ residual block}}}%
      \put(3920,120){\makebox(0,0){\strut{}\shortstack{ \vspace{8pt} \\ After last \\ residual block}}}%
    }%
    \gplgaddtomacro\gplfronttext{%
      \csname LTb\endcsname%
      \put(96,1056){\rotatebox{-270}{\makebox(0,0){\strut{}Accuracy}}}%
      \put(2273,-36){\makebox(0,0){\strut{}}}%
      \csname LTb\endcsname%
      \put(3536,1730){\makebox(0,0)[r]{\strut{}ROI-Pooling}}%
      \csname LTb\endcsname%
      \put(3536,1573){\makebox(0,0)[r]{\strut{}No bounding boxes}}%
    }%
    \gplbacktext
    \put(0,0){\includegraphics{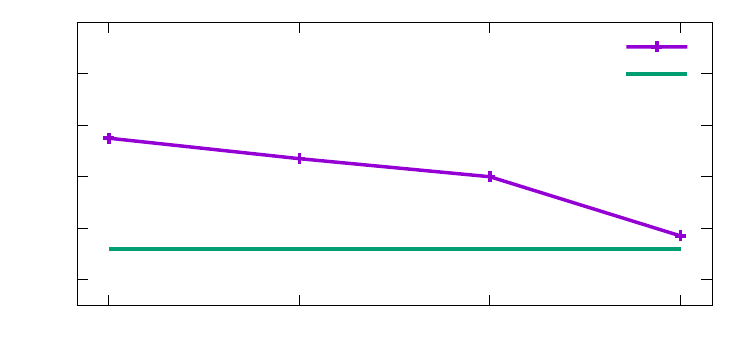}}%
    \gplfronttext
  \end{picture}%
\endgroup

%% file: kmax.tex
\begingroup
  \makeatletter
  \providecommand\color[2][]{%
    \GenericError{(gnuplot) \space\space\space\@spaces}{%
      Package color not loaded in conjunction with
      terminal option `colourtext'%
    }{See the gnuplot documentation for explanation.%
    }{Either use 'blacktext' in gnuplot or load the package
      color.sty in LaTeX.}%
    \renewcommand\color[2][]{}%
  }%
  \providecommand\includegraphics[2][]{%
    \GenericError{(gnuplot) \space\space\space\@spaces}{%
      Package graphicx or graphics not loaded%
    }{See the gnuplot documentation for explanation.%
    }{The gnuplot epslatex terminal needs graphicx.sty or graphics.sty.}%
    \renewcommand\includegraphics[2][]{}%
  }%
  \providecommand\rotatebox[2]{#2}%
  \@ifundefined{ifGPcolor}{%
    \newif\ifGPcolor
    \GPcolortrue
  }{}%
  \@ifundefined{ifGPblacktext}{%
    \newif\ifGPblacktext
    \GPblacktexttrue
  }{}%
  \let\gplgaddtomacro\g@addto@macro
  \gdef\gplbacktext{}%
  \gdef\gplfronttext{}%
  \makeatother
  \ifGPblacktext
    \def\colorrgb#1{}%
    \def\colorgray#1{}%
  \else
    \ifGPcolor
      \def\colorrgb#1{\color[rgb]{#1}}%
      \def\colorgray#1{\color[gray]{#1}}%
      \expandafter\def\csname LTw\endcsname{\color{white}}%
      \expandafter\def\csname LTb\endcsname{\color{black}}%
      \expandafter\def\csname LTa\endcsname{\color{black}}%
      \expandafter\def\csname LT0\endcsname{\color[rgb]{1,0,0}}%
      \expandafter\def\csname LT1\endcsname{\color[rgb]{0,1,0}}%
      \expandafter\def\csname LT2\endcsname{\color[rgb]{0,0,1}}%
      \expandafter\def\csname LT3\endcsname{\color[rgb]{1,0,1}}%
      \expandafter\def\csname LT4\endcsname{\color[rgb]{0,1,1}}%
      \expandafter\def\csname LT5\endcsname{\color[rgb]{1,1,0}}%
      \expandafter\def\csname LT6\endcsname{\color[rgb]{0,0,0}}%
      \expandafter\def\csname LT7\endcsname{\color[rgb]{1,0.3,0}}%
      \expandafter\def\csname LT8\endcsname{\color[rgb]{0.5,0.5,0.5}}%
    \else
      \def\colorrgb#1{\color{black}}%
      \def\colorgray#1{\color[gray]{#1}}%
      \expandafter\def\csname LTw\endcsname{\color{white}}%
      \expandafter\def\csname LTb\endcsname{\color{black}}%
      \expandafter\def\csname LTa\endcsname{\color{black}}%
      \expandafter\def\csname LT0\endcsname{\color{black}}%
      \expandafter\def\csname LT1\endcsname{\color{black}}%
      \expandafter\def\csname LT2\endcsname{\color{black}}%
      \expandafter\def\csname LT3\endcsname{\color{black}}%
      \expandafter\def\csname LT4\endcsname{\color{black}}%
      \expandafter\def\csname LT5\endcsname{\color{black}}%
      \expandafter\def\csname LT6\endcsname{\color{black}}%
      \expandafter\def\csname LT7\endcsname{\color{black}}%
      \expandafter\def\csname LT8\endcsname{\color{black}}%
    \fi
  \fi
    \setlength{\unitlength}{0.0500bp}%
    \ifx\gptboxheight\undefined%
      \newlength{\gptboxheight}%
      \newlength{\gptboxwidth}%
      \newsavebox{\gptboxtext}%
    \fi%
    \setlength{\fboxrule}{0.5pt}%
    \setlength{\fboxsep}{1pt}%
\begin{picture}(4320.00,2016.00)%
    \gplgaddtomacro\gplbacktext{%
      \csname LTb\endcsname%
      \put(372,384){\makebox(0,0)[r]{\strut{}$79$}}%
      \put(372,632){\makebox(0,0)[r]{\strut{}$80$}}%
      \put(372,880){\makebox(0,0)[r]{\strut{}$81$}}%
      \put(372,1128){\makebox(0,0)[r]{\strut{}$82$}}%
      \put(372,1376){\makebox(0,0)[r]{\strut{}$83$}}%
      \put(372,1624){\makebox(0,0)[r]{\strut{}$84$}}%
      \put(372,1872){\makebox(0,0)[r]{\strut{}$85$}}%
      \put(705,264){\makebox(0,0){\strut{}1}}%
      \put(1228,264){\makebox(0,0){\strut{}2}}%
      \put(1751,264){\makebox(0,0){\strut{}4}}%
      \put(2274,264){\makebox(0,0){\strut{}8}}%
      \put(2796,264){\makebox(0,0){\strut{}16}}%
      \put(3319,264){\makebox(0,0){\strut{}32}}%
      \put(3842,264){\makebox(0,0){\strut{}196}}%
    }%
    \gplgaddtomacro\gplfronttext{%
      \csname LTb\endcsname%
      \put(96,1128){\rotatebox{-270}{\makebox(0,0){\strut{}Accuracy}}}%
      \put(2273,84){\makebox(0,0){\strut{}K}}%
    }%
    \gplbacktext
    \put(0,0){\includegraphics{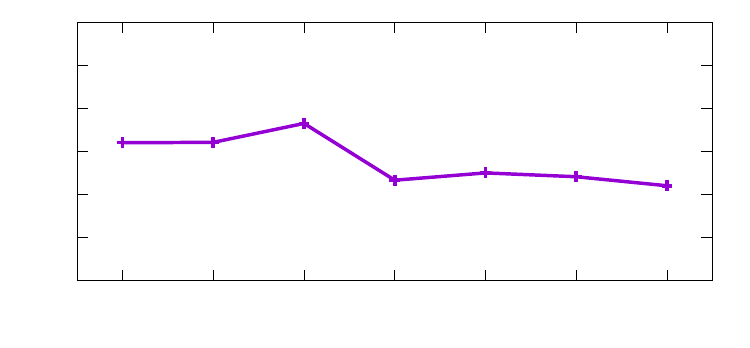}}%
    \gplfronttext
  \end{picture}%
\endgroup